\documentclass[journal]{IEEEtran}
\pdfoutput=1

\usepackage{color}

\usepackage{cite}

\usepackage[pdftex]{graphicx}
\graphicspath{{figures/}}

\usepackage{amsmath}
\usepackage{amssymb}

\usepackage{booktabs}
\usepackage{multirow}
\usepackage{stfloats}
\usepackage{tabularx}

\usepackage[caption=false,font=footnotesize]{subfig}

\usepackage{url}
\usepackage[hidelinks]{hyperref}

\begin{document}
	\title{Multi-Modal Interaction Graph Convolutional Network for Temporal Language Localization in Videos}
	
	\author{Zongmeng~Zhang,
		Xianjing~Han,
		Xuemeng~Song,
		Yan~Yan,
		and~Liqiang~Nie,~\IEEEmembership{Senior Member,~IEEE}}
	
	%



	\maketitle
	
	\begin{abstract}
		This paper focuses on tackling the problem of temporal language localization in videos, which aims to identify the start and end points of a moment described by a natural language sentence in an untrimmed video. However, it is non-trivial since it requires not only the comprehensive understanding of the video and sentence query, but also the accurate semantic correspondence capture between them. Existing efforts are mainly centered on exploring the sequential relation among video clips and query words to reason the video and sentence query, neglecting the other intra-modal relations (\textit{e.g.}, semantic similarity among video clips and syntactic dependency among the query words). Towards this end, in this work, we propose a Multi-modal Interaction Graph Convolutional Network (MIGCN), which jointly explores the complex intra-modal relations and inter-modal interactions residing in the video and sentence query to facilitate the understanding and semantic correspondence capture of the video and sentence query. In addition, we devise an adaptive context-aware localization method, where the context information is taken into the candidate moments and the multi-scale fully connected layers are designed to rank and adjust the boundary of the generated coarse candidate moments with different lengths. Extensive experiments on Charades-STA and ActivityNet datasets demonstrate the promising performance and superior efficiency of our model.
	\end{abstract}
	
	\begin{IEEEkeywords}
		Temporal Language Localization, Graph Convolutional Network, Video and Language.
	\end{IEEEkeywords}

	%
	\IEEEpeerreviewmaketitle

	\section{Introduction}
	\IEEEPARstart{I}{n} recent years, the flourish of multimedia devices has promoted the unprecedented growth of videos in various domains, highlighting the necessity of automatic video processing. In particular, owing to its great potential in the security domain, especially the video surveillance, temporal action localization in videos that aims to identify the start and end points of an action query has attracted more and more researchers~\cite{TIP2018stalstm, TIP2019wta, Zeng_2019, CVPR2016Ma, CVPR2016singh, CVPR2009yuan, tip2020RAM, tip2011SLCHA}. Traditionally, the action queries are described by keywords from a pre-defined set, like running or jumping. Nevertheless, in real-world scenarios, long untrimmed videos usually involve a large number of objects and complex activities, which are hard to be pre-defined. In light of this, in this work, we focus on the task of temporal language localization in videos, where the activity query is described by the natural language.
	
	\begin{figure}[ht]
		\centering
		\includegraphics[width=0.99\linewidth]{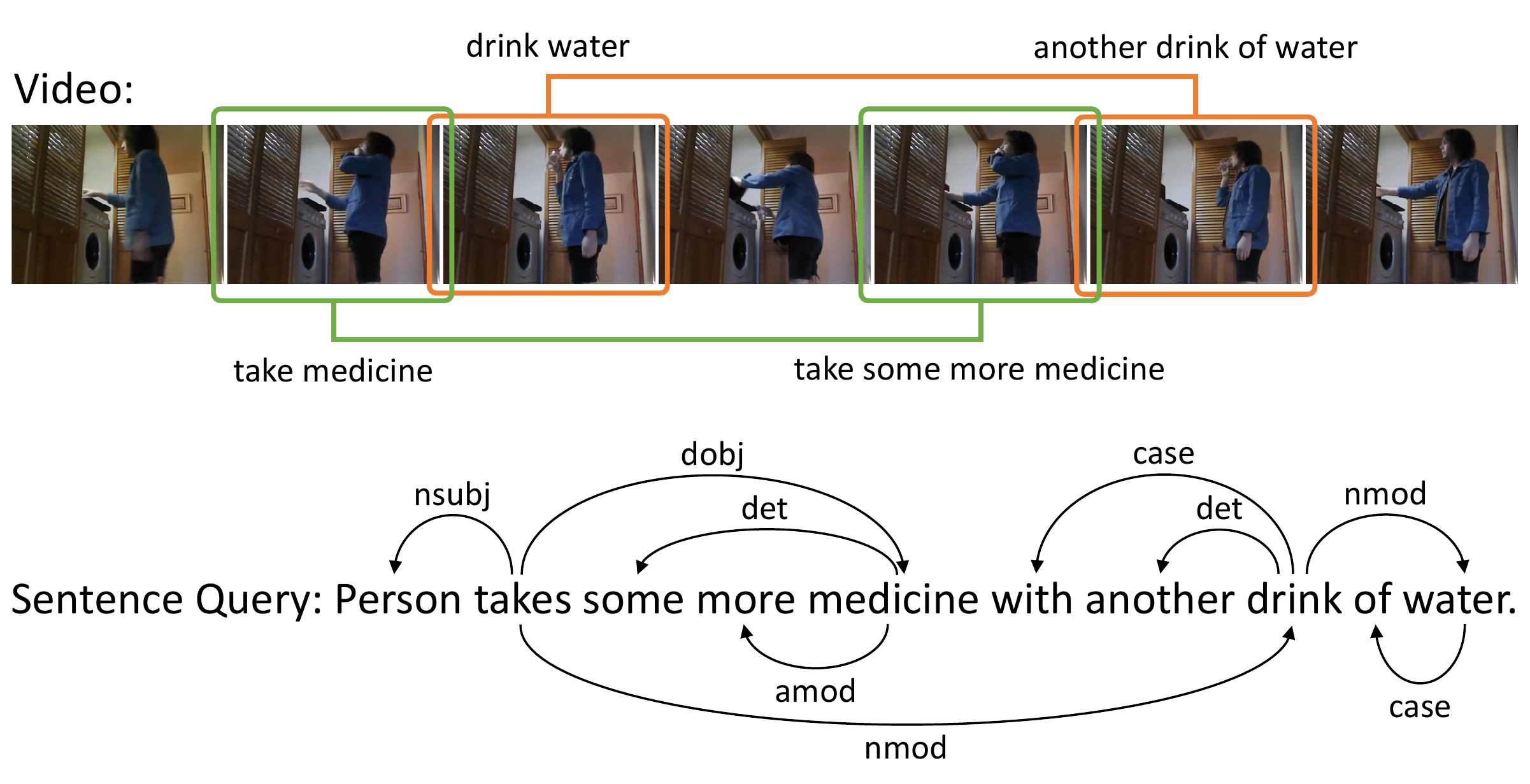}
		\caption{An example of the intra-modal relations in the video and sentence query. The clips in boxes of the same color present high semantic similarity, \textit{e.g.}, ``drink water'' and ``another drink of water''. The arrows between words show their syntactic dependencies, \textit{e.g.}, ``another'' is the determiner of ``drink''. In a sense, these relations can facilitate the video reasoning and strengthen the action understanding of the sentence query, and hence promote the moment localization for the query ``Person takes some more medicine with another drink of water''.}
		\label{intra-modal-relations}
	\end{figure}
	
	In fact, some research efforts have been dedicated to the task of temporal language localization in videos~\cite{gao2017tall, role, chen-etal-2018-temporally, He_2019, qspn_2019, sap, smrl2019, chen2019localizing, mithun2019weakly}. Since the semantic correspondence capture between the video and sentence query plays a pivotal role in this context, earlier studies~\cite{gao2017tall, Hendricks_2017, acrn, Jiang_2019, tripnet} mainly adopt the sliding window strategy to generate dense candidate moments and then explore the inter-modal interactions between the candidate moments and sentence query with various attention mechanisms~\cite{xu2015attention, coattention2016, chaplot2018gated}. Though these methods have achieved promising performance, they mainly focus on the inter-modal interactions, ignoring the sequential dependencies among video clips or query words, which are also crucial to the understanding of the video or sentence query. Towards this end, some efforts~\cite{chen-etal-2018-temporally, Yuan_2019, qspn_2019, Ghosh2019ExCLEC} have been made to capture the sequential relation among the video clips and query words with the recurrent neural networks (RNN)~\cite{lstm}. In fact, besides the sequential relation, there also exist other intra-modal relations in the video and sentence query, such as the semantic similarity among video clips and syntactic dependency among the query words, as shown in Figure~\ref{intra-modal-relations}. These relations are essential for the comprehensive understanding of the video and sentence query, yet they are overlooked by existing methods as it is challenging to capture these relations via a sequential manner. In addition, to facilitate the flexible moment localization, where the target moment length is unfixed, existing methods~\cite{gao2017tall, Hendricks_2017, chen-etal-2018-temporally, Ge_2019} mainly utilize the pooling or sampling strategy to regularize the representations of candidate moments. Nevertheless, the pooling or sampling strategy tends to merely retain partial prominent information of the candidate moment, and hence they inevitably suffer from the enormous information loss regarding the candidate moments, resulting in the suboptimal performance.
	
\begin{figure*}[ht]
		\centering
		\includegraphics[width=0.96\linewidth]{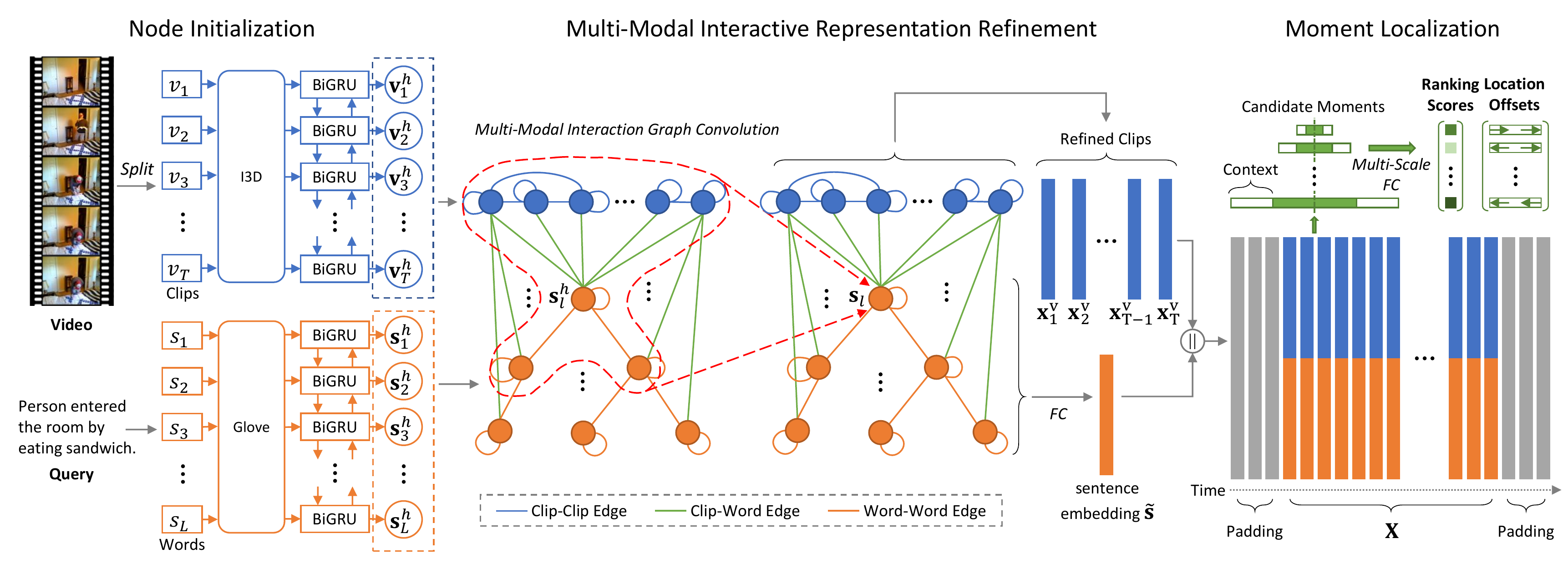}
		\caption{Pipeline of the proposed Multi-Modal Interaction Graph Convolutional Network (MIGCN). We first employ BiGRU to embed the clip and word features as the initial node representations of the multi-modal interaction graph. Then we utilize the graph convolution to refine the representation of the clip and word nodes with both the intra-modal relation and inter-modal interaction modeling. Based on the multi-modal fused clip representation matrix $\mathbf{X}$, we use sliding windows to derive a set of coarse candidate moments with different lengths. Ultimately, we introduce the adaptive context-aware moment localization module with multi-scale fully connected layers to predict the ranking scores and location offsets of each candidate moment. The red dashed line represents the information aggregation for the node $\mathbf{s}_l^h$ by the graph convolution process.}
		\label{model-pipeline}
	\end{figure*}

	To address the aforementioned issues, in this work, we devise a multi-modal interaction graph convolutional network (MIGCN) for temporal language localization. As shown in Figure~\ref{model-pipeline}, both the intra-modal relations and inter-modal interactions residing in the video and sentence query are comprehensively explored by the graph convolutional network (GCN)~\cite{kipf2017semi} which has been proven to be effective in propagating information among data with complex relations~\cite{NIPS2017_7231, Yan2018SpatialTG, Yao_2019, video_relation_mm2019, gcnNVS, gcnVSOD, gcnLGN, gcnZSD}. In particular, we first adopt BiGRU~\cite{bigru} to encode the video and sentence query, where we split the video into several clips without overlapping to reduce the computational complexity. To promote the representation learning of the video and sentence query by jointly modeling the intra-modal relations and inter-modal interactions, we then introduce the multi-modal interaction graph, comprising two types of nodes (clip node and word node), and edges that compile intra-modal relations and inter-modal interactions among the clips and words. Specifically, we incorporate the temporally adjacent relation and semantic correlation among video clips, and the syntactic dependency among words as the intra-modal relations, taking semantic correspondence between the video and sentence query as the inter-modal interactions. Based on this graph, we utilize the graph convolution to fulfil both the intra- and inter-modal refinement over the node representation learning. To facilitate the target moment with flexible lengths, we employ sliding windows with different sizes to generate a set of coarse candidate moments with different lengths. As for the ranking and boundary adjustment of these coarse candidate moments, we devise an adaptive context-aware localization method, where the context information is considered to learn the ranking scores and boundary offsets of the coarse candidate moments with less information loss through the multi-scale fully connected layers.
	
	The main contributions of this work can be summarized in threefold:
	\begin{itemize}
		\item We propose a multi-modal interaction graph convolutional neural network, where a graph comprising both video clip nodes and word nodes is constructed to jointly explore the complex intra-modal relations and inter-modal interactions residing in the video and sentence query. To the best of our knowledge, this is the first attempt to construct a multi-modal graph to tackle the problem of temporal language localization in videos.
		
		\item We devise an adaptive context-aware localization method, which employs the multi-scale fully connected layers and considers the context information, to rank the variable-length candidate moments with less information loss and promote the accurate candidate moment boundary adjustment.
		
		\item Extensive experiment results demonstrate the promising performance and efficiency of our MIGCN, compared with the state-of-the-art methods on two large datasets, Charades-STA~\cite{gao2017tall} and ActivityNet~\cite{krishna2017dense}. As a byproduct, we have released the codes and involved parameters to benefit other researchers\footnote{\url{https://github.com/zmzhang2000/MIGCN/}.}.
	\end{itemize}
	
	\section{Related Work}
	\subsection{Temporal Language Localization in Videos}
	The task of temporal language localization in videos is to determine the start and end points in an untrimmed video regarding the activity described in a sentence query, which was first introduced by Gao~\textit{et al.}~\cite{gao2017tall} and Hendricks~\textit{et al.}~\cite{Hendricks_2017}. The method in \cite{gao2017tall} concatenates the representations of the candidate moment and sentence query to estimate the moment-sentence alignment score and introduces temporal regression for the moment boundary adjustment, while~\cite{Hendricks_2017} targets at measuring the similarity between the candidate moment and sentence query in a semantic space. Given that the semantic correlation capture of the given video and sentence query constitutes a pivotal part in this task, the researchers have resorted to various attention mechanisms to enhance the interaction modeling between the video and sentence query. For example, Liu~\textit{et al.}~\cite{acrn} developed a memory attention mechanism to emphasize the visual features that are highly correlated to the sentence query. It is worth noting that the intra-modal relations in video and sentence query are crucial to the representation learning, yet they are overlooked by these methods. In order to tackle this issue, Yuan~\textit{et al.}~\cite{Yuan_2019} encoded the sequential relations via Bi-directional LSTM to generate representations of the video and sentence query with the sequential contextual information. In fact, besides the temporal relations, there exist other intra-modal relations in the video and sentence query, \textit{e.g.}, the semantic similarity among video clips and syntactic dependency among query words, which can strengthen the representation learning. Towards this end, we targeted at comprehensively explore the intra-modal relations and inter-modal interactions residing in the video and sentence query to boost the model performance on the temporal language localization task.
	
	\subsection{Graph Convolutional Networks}
	In recent years, graph convolutional networks (GNN) have drawn increasing attention due to their successful applications in various tasks~\cite{NIPS2017_7231, Yan2018SpatialTG, Yao_2019, video_relation_mm2019, gcnNVS, gcnVSOD, gnnMVGNN, gcnLGN, gnnA2GNN, gcnZSD, Xu2019GTADSL}. At the beginning, Scarselli~\textit{et al.}~\cite{GNN2009} introduced the graph neural network for graph-focused and node-focused applications by extending recursive neural networks and random walk models. Inspired by this, Kipf~\textit{et al.}~\cite{kipf2017semi} presented the graph convolutional network (GCN) and defined the convolution on non-grid structures. 
	
	Since then, GCN has been employed in various tasks. For example, in natural language processing, Marcheggiani and Titov~\cite{Marcheggiani_2017} employed GCN to model the syntactic dependency among words in sentence and learn the latent representations of words. In video understanding, Zhang~\textit{et al.}~\cite{TIP2020trg} put forward a temporal reasoning graph, which captures the temporal relation among video frames, to tackle the task of action recognition. Moreover, in the context of temporal language localization, Zeng~\textit{et al.}~\cite{Zeng_2019} introduced an action proposal graph to model the relations among different proposals, while Zhang~\textit{et al.}~\cite{zhang2019man} presented an iterative graph adjustment network to exploit the graph-structured moment relations. Although these efforts have achieved promising performance, they only focus on enhancing the single-modal representation learning with GCN, while overlooking the potential of GCN in propagating information on multi-modal relations residing in the video and sentence query.
	
	In fact, there have been several multi-modal based works employing GCN to promote the inter-modal interaction and achieving improved performance. For example, in the task of object grounding with textual description, Chen~\textit{et al.}~\cite{chen2019object} utilized the GCN to enhance the reasoning of the object and text, as well as boost the information passing among the object and text modalities. In addition, Bajaj~\textit{et al.}~\cite{bajaj2019g3raphground} proposed a grounding architecture with three connected graphs to tackle the language grounding, where a phrase graph and a visual graph are designed to boost the intra-modal representation, and then based on that a fusion graph is derived to enhance the inter-modal interaction. In this work, we focused on promoting both the video and sentence representation learning via the multi-modal interaction graph to tackle the task of temporal language localization in videos. In addition, it is worth noting that different from the graphs with entirely learned edges in other vision-language tasks~\cite{chen2019object, bajaj2019g3raphground, li2019relation}, our proposed multi-modal interaction graph explicitly leverages the temporally adjacent relation and semantic similarity among video clips, and the syntactic dependency in sentence by predefined adjacency matrices. Compared with other multi-modal graph methods, the pre-established adjacency matrices in our method are more interpretable and able to impose more powerful inductive biases to the model. 
	

	\section{Method}
	\subsection{Problem Formulation}
	In this work, we aim to tackle the problem of temporal language localization in videos. Suppose we have an untrimmed video $V=\{v_t\}_{t=1}^T$, which can be split into $T$ parts, where $v_t$ denotes the $t$-th video clip, and $T$ is the number of video clips. Besides, we have a sentence query $S=\{s_l\}_{l=1}^L$, where $s_l$ is the $l$-th word in the sentence and $L$ represents the sentence length. For each sentence query, we have a ground truth start-end points of the target moment represented as $(\tau^s, \tau^e)$. In a sense, our goal is to learn a mapping function $\mathcal{F}$ defined as follows:
	\begin{equation}
	\mathcal{F}: (V, S) \to (\tau^s, \tau^e).
	\end{equation}

	\subsection{Multi-Modal Interaction Graph Construction}
	Undoubtedly, it is essential for the temporal language localization task to comprehensively reason the given video and sentence. Existing efforts~\cite{gao2017tall, Hendricks_2017, role, Yuan_2019, Jiang_2019} mainly employ the sequential structures (\textit{e.g.}, recurrent neural networks) to reason the video and sentence due to their intrinsic sequential property. Nevertheless, apart from the sequential dependency, there are also other intra-model relations that can facilitate the reasoning of the video and sentence, \textit{e.g.}, the semantic similarity among the video clips and syntactic dependency among the sentence words. To boost these intra-model relations learning, we resort to the graph neural network, where these relations can be explicitly modeled by the edges in the graph rather than implicitly excavated from training data. These edges in graph connects correlated clips and words, which impose more powerful inductive biases~\cite{battaglia2018relational} (\textit{i.e.}, more prior knowledge) on the neural network, and thus ease the relation learning. As the key of the temporal language localization task is to capture the semantic correspondence between the video clips and sentence query, in addition to the aforementioned intra-modal relations, we also model the inter-modal interaction among the video and sentence, and hence introduce a multi-modal interaction graph.
	
	\noindent \textbf{Node Initialization.}
	The multi-modal interaction graph has two types of nodes: clip nodes and word nodes. To initialize the clip node, we employ the pre-trained model~\cite{i3d, two-stream, c3d} to extract the visual feature $\mathbf{v}_t$ of the $t$-th clip. Then, to comprehensively explore the semantic information in the clip sequence, we employ BiGRU (bi-directional GRU network) to encode the whole video. In particular, the BiGRU network comprises a $\overrightarrow{GRU}^v$ moving forward from the start to the end of the video, a $\overleftarrow{GRU}^v$ moving in the opposite direction, and a fully connected layer $f^v$, which takes the concatenation of the $t$-th hidden states of the two GRUs as the input. Ultimately, the output is used as the initial clip node representation $\mathbf{v}_t^h$, which can be formulated as follows: 
	\begin{equation}
	\left\{
	\begin{aligned}
	\overrightarrow{\mathbf{h}}_t^{v} &= \overrightarrow{GRU}^v(\mathbf{v}_t, \overrightarrow{\mathbf{h}}_{t-1}^{v}),\\
	\overleftarrow{\mathbf{h}}_t^{v} &= \overleftarrow{GRU}^v(\mathbf{v}_t, \overleftarrow{\mathbf{h}}_{t+1}^{v}),\\
	\mathbf{v}_t^h &= f^v(\overrightarrow{\mathbf{h}}_t^{v} \| \overleftarrow{\mathbf{h}}_t^{v}),
	\end{aligned}
	\right.
	\end{equation}
	where $\overrightarrow{\mathbf{h}}_t^{v}$ and $\overleftarrow{\mathbf{h}}_t^{v}$ denote the $t$-th clip hidden states of the forward and backward GRUs, respectively. $\|$ signifies the concatenation operation and the Leaky ReLU active function is adopted for $f^v$.
	
	The word node can be initialized in a similar manner. We first project each word $s_l$ into the embedding $\mathbf{s}_l$ by Glove~\cite{pennington2014glove} and then utilize BiGRU to encode each word with the context information of the sentence, which can be defined as follows, 
	\begin{equation}
	\left\{
	\begin{aligned}
	\overrightarrow{\mathbf{h}}_l^{s} &= \overrightarrow{GRU}^s(\mathbf{s}_l, \overrightarrow{\mathbf{h}}_{l-1}^{s}),\\
	\overleftarrow{\mathbf{h}}_l^{s} &= \overleftarrow{GRU}^s(\mathbf{s}_l, \overleftarrow{\mathbf{h}}_{l+1}^{s}),\\
	\mathbf{s}_l^h &= f^s(\overrightarrow{\mathbf{h}}_l^{s} \| \overleftarrow{\mathbf{h}}_l^{s}),
	\end{aligned}
	\right.
	\end{equation}
	where $\overrightarrow{\mathbf{h}}_l^{s}$ and $\overleftarrow{\mathbf{h}}_l^{s}$ are the $l$-th word hidden states of the forward and backward GRUs, respectively. $f^s$ is the fully connected layer with Leaky ReLU active function. $\mathbf{s}_l^h$ is the final initialization for the $l$-th word node.
	
	\noindent \textbf{Clip-Clip Edge.} 
	Intuitively, the sequence order of clips in a video reflects the temporally adjacent relation among these clips. Meanwhile, in the context of temporal language localization, the given sentence query may correspond to multiple clips over the whole video, where these clips tend to be visually similar and semantically correlated. Consequently, to explore both the temporal and semantic correlations and enhance the representation learning of clips, we devise two types of clip-clip edges. On the one hand, for the temporally adjacent clip pairs, we define a set of temporal correlated edges as follows:
	\begin{equation}
	\mathcal{E}^t = \{(v_i, v_{i+1})|i \in \{1, 2, \ldots, T-1\}\},\\
	\end{equation}
	where $(v_i, v_{i+1})$ represents the edge between the $i$-th and the next clip nodes. We set the weight of each edge $(v_i, v_{i+1})\in\mathcal{E}^t$ as $1$.
	On the other hand, to model the semantic correlation, we link two clip nodes if they share the similar visual content. In particular, we define a set of semantic correlated edges as follows:
	\begin{equation}
	\mathcal{E}^s = \{(v_i, v_j)|d_c(\mathbf{v}_i, \mathbf{v}_j) > \theta \wedge i\ne j\},
	\end{equation}
	where $ i,j \in \{1,2,\ldots,T\}$. $d_c(\mathbf{v}_i, \mathbf{v}_j)$ is the cosine similarity between the $i$-th and $j$-th clip features extracted by the pre-trained model, which can be computed by:
	\begin{equation}
	d_c(\mathbf{v}_i, \mathbf{v}_j) = \frac{\mathbf{v}_i^T\mathbf{v}_j}{\|\mathbf{v}_i\|_2 \cdot \|\mathbf{v}_j \|_2}.
	\end{equation}
	$\theta$ is the pre-defined threshold. We set the weight of the edge $(v_i, v_j)\in\mathcal{E}^s$ as $d_c(\mathbf{v}_i, \mathbf{v}_j)$. Finally, we can obtain the clip-clip edge set $\mathcal{E}^{vv}=\mathcal{E}^t\cup\mathcal{E}^s\cup\mathcal{E}^l$, where $\mathcal{E}^l$ is the set of self-loop edges utilized to maintain the information of the node itself. We set the weight of each self-loop edge as 1. 
	
	\noindent \textbf{Word-Word Edge}. Considering the success of the syntactic dependency graph in the query semantic understanding~\cite{Marcheggiani_2017}, we extract the syntactic dependency among words by Stanford CoreNLP~\cite{manning-etal-2014-stanford} and represent each dependency relation as an edge. Accordingly, we derive a set of word-word edges denoted as follows:
	\begin{equation}
	\mathcal{E}^{ss} = \{(s_i, s_j)| \langle i,j \rangle \in \Omega \vee i=j \},
	\end{equation}
	where $\Omega$ denotes the syntactic dependency relation set extracted from the sentence query, while $\langle i,j \rangle$ indicates that there are the syntactic dependency between the $i$-th and the $j$-th words in the sentence, $i,j \in \{1,2,\ldots,L\}$. The weight of each edge $(s_i, s_j)\in\mathcal{E}^{ss}$ is set as $1$. Notably, self-loop edges are also included in $\mathcal{E}^{ss}$ to preserve the information of the words themselves.
	
	\noindent \textbf{Clip-Word Edge.} To promote the information propagation among different modalities, apart form the intra-modal edges, we also bridge each clip and each word to constitute the inter-modal edges. In particular, we connect each clip with each word and get the clip-word edge set as follows:
	\begin{equation}
	\mathcal{E}^{vs} = \{(v_i, s_j)|i \in \{1, 2, \ldots, T\}, j \in \{1, 2, \ldots, L\} \}.
	\end{equation}
	Given that our goal is to localize the moment described by the sentence query, the inter-modal interactions are particularly essential. Different from the static weight setting for clip-clip edges and word-word edges, the weight for each clip-word edge $(v_i, s_j)$ is dynamically updated according to the node similarity defined as $d_s(v_i, s_j) = (\mathbf{v}_i^h)^T\mathbf{s}_i^h$.
	
	\subsection{Multi-Modal Interactive Representation Refinement}
	Over the constructed multi-modal graph, we adopt the graph convolution process which is capable of propagating information among nodes with complex relations to enhance the clip and word representation learning. Basically, the general graph convolution can be implemented as follows:
	\begin{equation}\label{eq9}
	\mathbf{H} = \sigma(\mathbf{A} \mathbf{X} \mathbf{W}),
	\end{equation}
	where $\mathbf{H}$ represents the hidden representations of the nodes, and $\mathbf{A}$ is the adjacency matrix. $\mathbf{X}$ denotes the input node features, and $\mathbf{W}$ is the to-be-learned weight matrix. $\sigma$ represents the non-linear operation. In a sense, each node representation can be refined according to the representations of itself and its adjacent nodes with a graph convolution operation. In our context, the graph convolution refinement over the clip and word representations consists of two aspects: intra- and inter-modal refinements.
	
	\noindent \textbf{Intra-Modal Refinement.} According to Eqn.($\ref{eq9}$), the intra-modal node representation refinement can be implemented by:
	\begin{equation}
	\left\{
	\begin{aligned}
	\tilde{\mathbf{V}} &= ReLU(\mathbf{A}_{vv} \mathbf{V} \mathbf{W}_{vv}),\\
	\tilde{\mathbf{S}} &= ReLU(\mathbf{A}_{ss} \mathbf{S} \mathbf{W}_{ss}), \label{intra-modal refinement}
	\end{aligned}
	\right.
	\end{equation}
	where $\mathbf{V} = [\mathbf{v}_1^h; \mathbf{v}_2^h; \ldots; \mathbf{v}_T^h]$ and $\mathbf{S} = [\mathbf{s}_1^h; \mathbf{s}_2^h; \ldots; \mathbf{s}_L^h]$. $\tilde{\mathbf{V}} \in \mathbb{R}^{T \times d}, \tilde{\mathbf{S}} \in \mathbb{R}^{L \times d}$ are the refined clip and word node representations, respectively. $\mathbf{W}_{vv} \in \mathbb{R}^{d \times d}$ and $\mathbf{W}_{ss} \in \mathbb{R}^{d \times d}$ are the learnable parameters. $ReLU$ denotes the leaky ReLU function. $\mathbf{A}_{vv} \in \mathbb{R}^{T \times T}$ represents the clip node adjacency matrix, which is constructed according to the clip-clip edge set $\mathcal{E}^{vv}$, while $\mathbf{A}_{ss} \in \mathbb{R}^{L \times L}$ is the word node adjacency matrix constructed with the word-word edge set $\mathcal{E}^{ss}$. Note that if $(v_i,v_j) \notin \mathcal{E}^{vv}$, we set the value of $\mathbf{A}_{vv}(i,j)$ as $0$. Similarly, if $(s_i,s_j) \notin \mathcal{E}^{ss}$, we set the value of $\mathbf{A}_{ss}(i,j)$ as $0$.
	
	\noindent \textbf{Inter-Modal Refinement.} 
	To fulfil the inter-modal refinement, one simple method is to update the node representations of each modality according to the inter-modal adjacent relation as follows:
	\begin{equation}
	\left\{
	\begin{aligned}
	\mathbf{X}^v &= ReLU(\mathbf{A}_{sv} \tilde{\mathbf{S}} \mathbf{W}_{sv}),\\
	\mathbf{X}^s &= ReLU(\mathbf{A}_{vs} \tilde{\mathbf{V}} \mathbf{W}_{vs}),
	\end{aligned}
	\right.
	\label{naive-inter}
	\end{equation}
	where $\mathbf{X}^{v}$ and $\mathbf{X}^{s}$ are the refined clip and word node representations, respectively. $\mathbf{A}_{sv}$ and $\mathbf{A}_{vs}$ are respectively the word-clip and clip-word adjacency matrices, both constructed based on the clip-word edge set $\mathcal{E}^{vs}$. $\mathbf{W}_{sv}$ and $\mathbf{W}_{vs}$ are the parameters to be learned. Apparently, this simple graph convolution operation only refers the node representations from another modality (\textit{e.g.}, the sentence words) to refine one modality (\textit{e.g.}, the video clips), totally ignoring the modality inherent information, which may hinder the thorough capture of the semantic interactions between the two modalities and hence hurt the performance. Therefore, to realize the comprehensive inter-modal refinement, we resort to the gate mechanism (\textit{i.e.}, gated graph convolution~\cite{chen2019object}), and the inter-modal refinement over the clip node representation is then formulated as follows:
	\begin{equation}
	\left\{
	\begin{aligned}
	&\mathbf{H}_s = \mathbf{A}_{sv} \tilde{\mathbf{S}} \mathbf{W}_{sv}, \\
	&\mathbf{Z}_v = sigmoid([\tilde{\mathbf{V}} \| \mathbf{H}_s]\mathbf{W}_{gate,v}), \\
	&\mathbf{X}^{v} = ReLU(\mathbf{Z}_v \circ \tilde{\mathbf{V}} + (1 - \mathbf{Z}_v) \circ \mathbf{H}_s),
	\end{aligned}
	\right.
	\end{equation}
	where $\circ$ denotes the Hadamard multiplication of two matrices. $\mathbf{H}_s \in \mathbb{R}^{T \times d}$ is the word information obtained from the word node via the graph convolution, and $\mathbf{Z}_v \in \mathbb{R}^{T \times d}$ denotes the retain ratio matrix of clip representations. $\mathbf{W}_{sv} \in \mathbb{R}^{d \times d}$ and $\mathbf{W}_{gate,v} \in \mathbb{R}^{2d \times d}$ are parameters to be learned. $\mathbf{A}_{sv} \in \mathbb{R}^{T \times L}$ is the word-clip adjacency matrix constructed according to the clip-word edge set $\mathcal{E}^{vs}$\footnote{To keep $\mathbf{H}_s$ and $\tilde{\mathbf{S}}$ in the same scale and balance the impact of $\tilde{\mathbf{V}}$ and $\tilde{\mathbf{S}}$ on $\mathbf{X}^{v}$, we perform the row-wise normalization on $\mathbf{A}_{sv}$ to facilitate the information propagation.}.
	
	Similarly, we summarize the inter-modal refinement over the word node representation as follows:
	\begin{equation}
	\left\{
	\begin{aligned}
	&\mathbf{H}_v = \mathbf{A}_{vs} \tilde{\mathbf{V}} \mathbf{W}_{vs}, \\
	&\mathbf{Z}_s = sigmoid([\tilde{\mathbf{S}} \| \mathbf{H}_v]\mathbf{W}_{gate,s}), \\
	&\mathbf{X}^{s} = ReLU(\mathbf{Z}_s \circ \tilde{\mathbf{S}} + (1 - \mathbf{Z}_s) \circ \mathbf{H}_v),
	\end{aligned}
	\right.
	\end{equation}
	where $\mathbf{H}_v \in \mathbb{R}^{L \times d}$ is the clip information obtained from the clip node via the graph convolution, and $\mathbf{Z}_s \in \mathbb{R}^{L \times d}$ denotes the retain ratio matrix of word representations. $\mathbf{W}_{vs} \in \mathbb{R}^{d \times d}$ and $\mathbf{W}_{gate,s} \in \mathbb{R}^{2d \times d}$ are parameters to be learned. $\mathbf{A}_{vs}\in \mathbb{R}^{L \times T}$ is the clip to word adjacency matrix.
	
	Ultimately, we can obtain the refined clip and word representations $\mathbf{X}^v=[\mathbf{x}_1^v, \mathbf{x}_2^v, \ldots,\mathbf{x}_T^v] \in \mathbb{R}^{T \times d}$ and $\mathbf{X}^s=[\mathbf{x}_1^s, \mathbf{x}_2^s, \ldots,\mathbf{x}_L^s] \in \mathbb{R}^{L \times d}$, which encode both the intra-modal relations and inter-modal semantic interactions.
	
	\subsection{Moment Localization}
	Based on the refined clip and word representations, we resort to sliding windows with different sizes to first generate a set of coarse candidate moments with different lengths to facilitate the target moment localization with flexible lengths. To accurately localize the target moment, we propose an adaptive context-aware localization method to rank the candidate moments and adjust their boundaries, where the context information of the candidate moments is taken into account and the multi-scale fully connected layers are devised to adaptively tackle candidate moments with different lengths. 
	
	\noindent \textbf{Candidate Moment Representation.} We first generate variable-length candidate moments by a set of sliding windows with sizes of $\{\omega_m\}_{m=1}^M$. In particular, we slide the window with each size $\omega_m$ on the $T$ video clips with a stride of $\delta$ to generate candidate moments. Be aware that we discard the candidate moment whose boundary (either the start point or the end point) exceeds the video clip range. We denote the set of candidate moments generated by $\omega_m$ as $\{(t_c^{s,m}, t_c^{e,m})\}_{c=1}^{C_m}$, where $t_c^{s,m}$ and $t_c^{e,m}$ are respectively the start and end points of the $c$-th candidate moment generated by the sliding window with size $\omega_m$. In addition, considering that the information near the candidate moment is essential for adjusting the moment boundary, different from~\cite{chen-etal-2018-temporally, zhang2019cross}, we take the context information into account to get more accurate boundary offsets of candidate moments. In particular, we expand both the start and end points of the $c$-th candidate moment, which is formulated as:
	\begin{equation}
	\left\{
	\begin{aligned}
	\tilde{t}_{c}^{s,m} &= t_{c}^{s,m} - \frac{\omega_m}{2}, \\
	\tilde{t}_{c}^{e,m} &= t_{c}^{e,m} + \frac{\omega_m}{2}.
	\end{aligned}
	\right.
	\end{equation}
	In a sense, each candidate moment length is doubled by this context extension. In order to measure the matching degree between each candidate moment $(t_{c}^{s,m}, t_{c}^{e,m})$ and the sentence query, we first employ a non-linear transformation to get the sentence embedding $\tilde{\mathbf{s}} \in \mathbb{R}^{d}$ based on the refined word representations as follows:
	\begin{equation}
	\tilde{\mathbf{s}} = ReLU(\mathbf{W}_s\mathbf{X}^s),
	\end{equation}
	where $\mathbf{W}_s \in \mathbb{R}^{1 \times L}$ is the transformation parameter. Thereafter, we concatenate the sentence embedding to each clip representation of the candidate moment, \textit{i.e.}, $\mathbf{x}_t^v$, where $v_t \in (\tilde{t}_{c}^{s,m}, \tilde{t}_{c}^{e,m})$ and then let derive the moment representation as $\mathbf{X}_c^m \in \mathbb{R}^{2\omega_m \times 2d}$. Note that we pad zero to $\mathbf{X}_c^m$ if the extended start point or end point exceeds the video clip range, \textit{i.e.}, $[1, T]$.
	
	\noindent \textbf{Adaptive Context-Aware Localization.} Based on the candidate moment set, the task of temporal language localization can be converted to the candidate moment ranking problem. Similar to the previous work~\cite{gao2017tall}, apart from the candidate moment ranking, we also predict the offsets of the start and end point of the candidate moment with a sibling output layer to refine the location of candidate moment. Towards this end, existing methods~\cite{gao2017tall, Hendricks_2017, chen-etal-2018-temporally, Ge_2019} mainly employ the pooling or sampling strategy to unify the representation dimensions of candidate moments with different lengths. Nevertheless, the pooling or sampling strategy mainly focuses on retaining the prominent information of the candidate moment, which inevitably suffers from the information loss to some extent. Towards this end, to retain the information as much as possible, we devise the multi-scale fully connected layers to adaptively process the candidate moments with different lengths. In particular, we calculate the ranking score $r_c^m$ and offset $(d_c^{s,m}, d_c^{e,m})$ of the candidate moment $(t_{c}^{s,m}, t_{c}^{e,m})$ as follows:
	\begin{equation}\label{eq17}
	\left\{
	\begin{aligned}
	r_c^m &= sum(\mathbf{X}_c^m \circ \mathbf{W}^m_{r}) + b^m_{r}, \\
	d_c^{s,m} &= sum(\mathbf{X}_c^m \circ \mathbf{W}^m_{s}) + b^m_{s}, \\
	d_c^{e,m} &= sum(\mathbf{X}_c^m \circ \mathbf{W}^m_{e}) + b^m_{e},
	\end{aligned}
	\right.
	\end{equation}
	where $sum()$ is the sum operator over all elements in a matrix. $\mathbf{W}^m_{r}, \mathbf{W}^m_{s}$ and $\mathbf{W}^m_{e}$ are learnable weight matrices of the multi-scale fully connected layers, which have the same dimensions with the candidate moment representation $\mathbf{X}_c^m$. $b^m_{r}, b^m_{s}$ and $b^m_{e}$ are the corresponding biases. It is worth noting that as the dimension of $\mathbf{X}_c^m$ is decided by the sliding window size, all the candidate moments generated by a specific window would share the identical parameters in the multi-scale fully connected layers. Ultimately, the rectified boundary $(\hat{\tau}_c^{s,m}, \hat{\tau}_c^{e,m})$ is computed by:
	\begin{equation}
	\left\{
	\begin{aligned}
	\hat{\tau}_{c}^{s,m} &= t_{c}^{s,m} + d_{c}^{s,m}, \\
	\hat{\tau}_{c}^{e,m} &= t_{c}^{e,m} + d_{c}^{e,m}.
	\end{aligned}
	\right.
	\end{equation}
	
	\subsection{Learning}
	As for the optimization, we utilize the alignment loss~\cite{zhang2019cross}, ranking loss~\cite{chen2019localizing} and regression loss~\cite{gao2017tall}, where the former loss is used for encouraging the model to evaluate the alignment degree between each candidate moment and the ground truth moment, while the latter two losses are used for adjusting the ranking scores and boundaries of candidate moments.
	
	\noindent \textbf{Alignment Loss.} Similar with~\cite{zhang2019cross}, we employ the alignment loss to assign high ranking scores to the candidate moments in line with the ground truth moment and lower ranking scores to the other moments. In this work, we employ Intersection over Union (IoU) between each candidate moment $(t_c^{s,m}, t_c^{e,m})$ and the ground truth moment$({\tau}^s, {\tau}^e)$, denoted as $\gamma_c^m$, to represent their alignment degree.
	
	Accordingly, we utilize binary cross entropy and formulate the alignment loss as follows:
	\begin{equation}
	\left\{
	\begin{aligned}
	\hat{r}_c^m &= sigmoid(r_c^m), \\
	\mathcal{L}_{aln} &= -\frac{1}{C} \sum_{m=1}^{M} \sum_{c=1}^{C_m} \gamma_c^m\log(\hat{r}_c^m) + (1-\gamma_c^m)\log(1-\hat{r}_c^m),
	\end{aligned}
	\right.
	\end{equation}
	where $\hat{r}_c^m$ is the normalized ranking score, which is regarded as the predicted IoU of the $c$-th candidate moment generated by the sliding window with size $\omega_m$. $C_m$ is the number of the candidate moments generated by the sliding window with size $\omega_m$, and $C$ is the total number of the candidate moments. In particular, we set $\gamma_c^m$ to 0, if $\gamma_c^m$ is less than a pre-defined threshold $\lambda$, to promote the identification of the high-score candidate moment~\cite{zhang2019cross}.
	
	\noindent \textbf{Ranking Loss.} As there can be plenty of candidate moments having similar locations and lengths with the target moment, it is hard to select the optimal candidate moment among them. Therefore, to promote the model to distinguish the optimal candidate moment, similar with~\cite{chen2019localizing}, we adopt the multi-class cross entropy loss as the ranking loss, which can be formulated as follows:
	\begin{equation}
	\left\{
	\begin{aligned}
	c^*, m^* &= \mathop{\arg\max}_{c, m}{\gamma_c^m}, \\
	\mathcal{L}_{rank} &= -\log(\frac{\exp({r_{c^*}^{m^*}})}{\sum_{m=1}^{M} \sum_{c}^{C_m} \exp(r_c^m)}),
	\end{aligned}
	\right.
	\end{equation}
	where $r_{c^*}^{m^*}$ is the ranking score of the optimal candidate moment, which has the highest IoU with the ground truth moment. The ranking loss promotes the model to distinguish the optimal candidate moment from other candidates by raising its ranking scores while decreasing that of others.
	
	\noindent \textbf{Regression Loss.} Considering that the candidate moments generated by sliding windows with fixed lengths may not exactly align with the ground truth moment, we propose to predict the candidate moment offset to adjust the boundary and employ the offset regression presented in~\cite{gao2017tall} to optimize the location of candidate moment. The regression loss is formulated as:
	\begin{equation}
	\mathcal{L}_{reg} = SL_1(\tau^s - \hat{\tau}_{c^*}^{s,m^*})+SL_1(\tau^e - \hat{\tau}_{c^*}^{e,m^*}),
	\end{equation}
	where $SL_1()$ is the Smooth $L_1$ function~\cite{fastRCNN}. $(\hat{\tau}_{c^*}^{s,m^*}, \hat{\tau}_{c^*}^{e,m^*})$ is the rectified boundary of the candidate moment that has the highest IoU with the ground truth moment.
	
	Hence we obtain the final objective function of the proposed model as follows: 
	\begin{equation}
	\mathcal{L} = \mathcal{L}_{aln} + \alpha\mathcal{L}_{rank} + \beta\mathcal{L}_{reg},
	\end{equation}
	where $\alpha$ and $\beta$ are hyper-parameters to balance these items.
	
	\section{Experiments}
	\subsection{Datasets}
	To evaluate the proposed method, we conducted experiments on two benchmark datasets.
	
	\noindent \textbf{Charades-STA}~\cite{gao2017tall}: The Charades-STA dataset is built based on the Charades~\cite{sigurdsson2016hollywood} dataset, comprising $6,672$ videos of indoor activities. Totally, there are $12,408$ moment-sentence pairs in the training set and $3,720$ pairs in the testing set. The videos in Charades-STA are $30$ seconds on average and the annotated moments are $8$ seconds on average.
	
	\noindent \textbf{ActivityNet}~\cite{krishna2017dense}: We evaluate our model on another benchmark ActivityNet Captions built upon the videos from ActivityNet~\cite{actnet}, to demonstrate the robustness of the proposed model. The dataset contains $20,000$ videos, $37,421$ moment-sentence pairs for training and $34,536$ pairs for testing. On average, the videos in ActivityNet are $180$ seconds and each annotated moment lasts for $36$ seconds. 
	
	\subsection{Experimental Settings}
	\noindent \textbf{Evaluation Metric.} Similar with~\cite{gao2017tall}, we adopted the metric of ``R@$1$, IoU@$n$'' to measure our model. Specifically, this metric represents the percentage of the top one candidate moments for all sentence queries with IoU larger than $n$ for all the given sentence queries, where the IoU is calculated with the boundaries of the candidate moment and the ground truth moment.
	
	\begin{table*}[ht]
		\centering
		\caption{Performance Comparison on Charades-STA and ActivityNet Datasets in Terms of R@1, IoU@$n$ (\%). The Results of Other Methods are Reported According to Their Papers or Existing Reimplement.}
		\label{BaselineComparison}
		\begin{tabularx}{0.9\linewidth}{X<{\centering}|X<{\centering}|X<{\centering}|X<{\centering}|X<{\centering}|X<{\centering}}
			\toprule
			\multicolumn{2}{c|}{} & \multicolumn{2}{c|}{Charades-STA~\cite{gao2017tall}} & \multicolumn{2}{c}{ActivityNet~\cite{krishna2017dense}} \\
			\hline
			Feature & Method & R@1, IoU@0.5 & R@1, IoU@0.7 & R@1, IoU@0.3 & R@1, IoU@0.5 \\
			\hline
			\multirow{14}*{C3D} & MCN~\cite{Hendricks_2017} & 17.46 & 8.01 & 21.37 & 9.58 \\
			& CTRL~\cite{gao2017tall} & 23.63 & 8.89 & 28.70 & 14.00 \\
			& ACRN~\cite{acrn} & - & - & 31.29 & 16.17 \\
			& TGN~\cite{chen-etal-2018-temporally} & - & - & 43.81 & 27.93 \\
			& ABLR~\cite{Yuan_2019} & 24.36 & 9.01 & 55.67 & 36.79 \\
			& SAP~\cite{sap} & 27.42 & 13.36 & - & -\\
			& QSPN~\cite{qspn_2019} & 35.60 & 15.80 & 45.30 & 27.70 \\ 
			& RWM~\cite{He_2019} & 36.70 & 13.74 & 53.00 & 36.90 \\
			& CBP~\cite{wang2020temporally} & 36.80 & 18.87 & 54.30 & 35.76 \\
			& TripNet~\cite{tripnet} & 38.29 & 16.07 & 48.42 & 32.19 \\
			& TSP-PRL~\cite{wu2020treestructured} & 37.39 & 17.69 & 56.08 & 38.76 \\
			& 2D-TAN~\cite{2DTAN_2020_AAAI} & - & - & 59.45 & 44.51 \\
			& DRN~\cite{zeng2020dense} & \textbf{45.40} & \textbf{26.40} & - & 43.97 \\
			& \textbf{MIGCN (Ours)} & 42.26 & 22.04 & \textbf{60.03} & \textbf{44.94}\\
			\hline
			\multirow{2}*{Two-Stream} & TSP-PRL~\cite{wu2020treestructured} & 45.30 & 24.73 & - & - \\
			& \textbf{MIGCN (Ours)} & \textbf{51.80} & \textbf{29.33} & - & - \\
			\hline
			\multirow{4}*{I3D} & ExCL~\cite{Ghosh2019ExCLEC} & 44.10 & 23.30 & - & - \\
			& MAN~\cite{zhang2019man} & 46.63 & 22.72 & - & -\\
			& DRN~\cite{zeng2020dense} & 53.09 & 31.75 & - & - \\
			& \textbf{MIGCN (Ours)} & \textbf{57.10} & \textbf{34.54} & - & -\\
			\bottomrule
		\end{tabularx}
	\end{table*}
	
	\noindent \textbf{Implementation Details.}
	For the video representation, we split each video in Charades-STA into $75$ clips and that in ActivityNet into $300$ clips. For Charades-STA, based on the previous studies, we utilized three mainstream frameworks: I3D network~\cite{i3d}, C3D network~\cite{c3d} and Two-Stream network~\cite{two-stream} to extract the visual features of $1,024$-D, $4,096$-D, and $8,192$-D, respectively. For ActivityNet, we directly utilized the publicly available 500-D C3D features\footnote{\url{http://activity-net.org/challenges/2016/download.html\#c3d/}.}, which are derived by PCA over the original $4,096$-D visual features extracted by C3D network. 
	
	For the sentence representation, we first tokenized the sentence query and extracted the syntactic dependency graph using Stanford CoreNLP~\cite{manning-etal-2014-stanford} toolkit. Then we employed the Glove~\cite{pennington2014glove} model pre-trained on Wikipedia to obtain the $300$-D embedding feature for each word token. The max length of the sentence query in Charades-STA and ActivityNet are respectively set to $10$ and $50$. We truncated the sentences exceeding the maximum length and padded the shorter ones with zeros.
	
	In the training phase, the batch size is set to $128$ and Adam optimizer is used for optimization. The learning rate is set to $0.001$ and $0.0003$ for Charades-STA and ActivityNet, respectively. Furthermore, we added a weight decay item with factor $0.00001$ and a dropout probability $0.5$ to improve the performance. The node dimension $d$ is set as $256$ and $512$ for Charades-STA and ActivityNet, respectively, while the dimension of GRU hidden state is half of the node dimension. The threshold $\theta$ and $\lambda$ are set to $0.7$ and $0.3$. The trade-off $\alpha$ and $\beta$ are set to $0.1$ and $0.001$ for Charades-STA, $1$ and $0.001$ for ActivityNet, respectively. We set $6$ window sizes of $[6, 12, 18, 24, 30, 36]$ for Charades-STA and $7$ window sizes of $[6, 12, 24, 48, 96, 192, 288]$ for ActivityNet. Window stride $\delta$ is set to $3$. More details are in our released code.
	
	\subsection{Comparison Among Methods}
	We compared the proposed MIGCN with other state-of-the-art methods on Charades-STA and ActivityNet. Among these methods, MCN~\cite{Hendricks_2017}, CTRL~\cite{gao2017tall}, ACRN~\cite{acrn}, TGN~\cite{chen-etal-2018-temporally}, SAP~\cite{sap}, QSPN~\cite{qspn_2019}, CBP~\cite{wang2020temporally}, MAN~\cite{zhang2019man}, 2D-TAN~\cite{2DTAN_2020_AAAI} and DRN~\cite{zeng2020dense} are based on the proposal generation and ranking. To improve the computation efficiency, ABLR~\cite{Yuan_2019} and ExCL~\cite{Ghosh2019ExCLEC} directly predict the location without the dense proposal generation. In addition, RWM~\cite{He_2019}, TripNet~\cite{tripnet}, and TSP-PRL~\cite{wu2020treestructured} regard the task as a sequential decision making process and employ the reinforcement learning paradigm to iteratively adjust the boundary of the moment. According to the original papers, ExCL~\cite{Ghosh2019ExCLEC}, MAN~\cite{zhang2019man}, and DRN~\cite{zeng2020dense} adopt the I3D features, TSP-PRL~\cite{wu2020treestructured} selects both the Two-Stream and C3D features, while DRN~\cite{zeng2020dense} and other methods use the C3D features.
	
	Table~\ref{BaselineComparison} shows the performance of different methods, from which we could observe that: 1) The proposed MIGCN exhibits superiority over other methods in most scenarios, demonstrating the effectiveness and robustness of our model. 2) On Charades-STA dataset, MIGCN is inferior to DRN with C3D feature but outperforms DRN with I3D feature. One possible reason is that the I3D network has higher temporal resolution and deeper architecture~\cite{i3d} than C3D and hence is able to learn the representation of activity with more concrete semantics. Compared with the feature pyramid construction in DRN, which may collapse the semantics in the I3D feature, the graph convolution in MIGCN is more likely to completely capture the feature semantics and thus result the better performance. 3) Both MIGCN and TSP-PRL perform better on the Two-Stream features than C3D features. One possible reason is that compared with the C3D features, the Two-Stream with optical flow features are more powerful in capturing temporal information and recognizing the actions in the clips and can benefit the temporal language localization task. Form the observation 2) and 3), we found that MIGCN exhibits different degrees of improvement over other methods with different visual features, but it is difficult to provide intuitive explanations. Therefore, we will dive into the effects of various visual and linguistic representations, \textit{e.g.}, VL-BERT~\cite{vlbert} and BERT~\cite{bert}, on vision-language tasks in our future work. And 4) MIGCN, MAN and 2D-TAN that have graph or graph-like components achieve better performance than most purely sequence-based  networks, \textit{e.g.}, CTRL, TGN and CBP. This could be an evidence that the graph architecture may have advantage in this task and benefit the performance in general.

	\begin{table}[b]
		\centering
		\caption{Performance Comparison Among the Proposed MIGCN and Its Derivations in Terms of R@1, IoU@$n$ (\%) Based on the I3D Features of Charades-STA and C3D Features of ActivityNet.}
		\label{AblationStudy}
		\begin{tabular}{c|c|c|c|c}
			\toprule
			\multicolumn{1}{c|}{} & \multicolumn{2}{c|}{Charades-STA~\cite{gao2017tall}} & \multicolumn{2}{c}{ActivityNet~\cite{krishna2017dense}} \\
			\hline
			\multirow{2}{*}{Method} & R@1, & R@1, & R@1, & R@1, \\
			& IoU@0.5 & IoU@0.7 & IoU@0.3 & IoU@0.5 \\
			\hline
			MIGCN-w/o-GCN & 42.26 & 24.27 & 54.21 & 37.46 \\
			MIGCN-G3G & 44.30 & 25.78 & 57.34 & 41.17 \\
			MIGCN-w/o-Inter & 43.15 & 24.81 & 59.03 & 41.96 \\
			MIGCN-w/o-Temp & 56.61 & 33.52 & 59.91 & 44.76 \\
			MIGCN-w/o-Seman & 53.68 & 32.31 & 57.76 & 42.38 \\
			MIGCN-w/o-Syntac & 55.19 & 33.66 & 59.71 & 43.99 \\
			MIGCN-w/o-Gate & 54.73 & 31.67 & 58.44 & 43.64 \\
			\hline
			MIGCN-MP & 36.21 & 16.32 & 43.50 & 27.06 \\
			MIGCN-Sample & 50.81 & 29.27 & 44.57 & 28.87 \\
			MIGCN-w/o-Context & \textbf{57.82} & 34.30 & 53.97 & 37.97 \\
			\hline
			\textbf{MIGCN} & 57.10 & \textbf{34.54} & \textbf{60.03} & \textbf{44.94} \\
			\bottomrule
		\end{tabular}
	\end{table}
	
	\subsection{Ablation Study}
	
	\begin{table*}[ht]
		\centering
		\caption{Efficiency Comparison Among Different Methods on Charades-STA and ActivityNet Dataset.}
		\label{Efficiency}
		\begin{tabularx}{0.9\linewidth}{X<{\hsize=0.6\hsize\centering}|X<{\hsize=1.6\hsize\centering}|X<{\centering}|X<{\hsize=0.9\hsize\centering}|X<{\centering}|X<{\centering}|X<{\centering}|X<{\hsize=0.9\hsize\centering}|X<{\centering}|X<{\centering}}
			\toprule
			\multicolumn{2}{c|}{} & \multicolumn{4}{c|}{Charades-STA~\cite{gao2017tall}} & \multicolumn{4}{c}{ActivityNet~\cite{krishna2017dense}} \\
			\hline
			Feature & Method & Time~$\downarrow$ & FPS~$\uparrow$ & FLOPs~$\downarrow$ & Params~$\downarrow$ & Time~$\downarrow$ & FPS~$\uparrow$ & FLOPs~$\downarrow$ & Params~$\downarrow$ \\
			\hline
			\multirow{6}*{C3D}& CTRL~\cite{gao2017tall} & 2541.06ms & 0.28K & 0.82G & 21.60M & - & - & - & - \\
			& MAC~\cite{Ge_2019} & 2141.36ms & 0.33K & \textbf{0.61G} & 23.10M & - & - & - & - \\
			& DRN~\cite{zeng2020dense} & 162.97ms & 4.31K & 4.05G & 35.66M & - & - & - & - \\
			& 2D-TAN~\cite{2DTAN_2020_AAAI} & - & - & - & - & 1228.91ms & 3.02K & 498.71G & 91.59M \\
			& TSP-PRL~\cite{wu2020treestructured} & \textbf{49.53ms} & \textbf{14.17K} & 139.81G & 182.46M & - & - & - & - \\
			& \textbf{MIGCN (Ours)} & 51.45ms & 13.64K & 3.88G & \textbf{10.94M} & \textbf{136.50ms} & \textbf{27.15K} & \textbf{1.18G} & \textbf{8.74M} \\
			\hline
			I3D & \textbf{MIGCN (Ours)} & \textbf{28.03ms} & \textbf{25.04K} & \textbf{0.32G} & \textbf{2.26M} & - & - & - & - \\
			\bottomrule
		\end{tabularx}
	\end{table*}
	
	\label{sec:AblationStudy}
	
	We conducted the ablation study to demonstrate the effects of the components in MIGCN.
	
	\noindent \textbf{Ablation on Multi-Modal Interaction Graph.} To exploit the effect of the multi-modal interaction graph convolution, we designed the following derivations of MIGCN:
	\begin{itemize}
		\item \textbf{MIGCN-w/o-GCN}: We removed the total process of multi-modal interactive representation refinement and generated the candidate moment representation merely with the output of BiGRU, \textit{i.e.}, $\mathbf{v}_t^h$ and $\mathbf{s}_l^h$.
		
		\item \textbf{MIGCN-G3G}: To verify the effectiveness of the multi-modal interaction graph, we replaced our graph refinement procedure with that in \textsc{G$^3$raphGround}~\cite{bajaj2019g3raphground}. MIGCN-G3G first executes intra-modal refinement to refine the video clip and sentence representations. Then MIGCN-G3G concatenates the sentence representations to each clip representation and conducts the graph convolution using clip-clip edges to fuse the two modalities.
		
		\item \textbf{MIGCN-w/o-Inter}: We disabled the inter-modal refinement and only used the intra-modal refinement in the graph convolution refinement process to validate the importance of the inter-modal interactions.
		
		\item \textbf{MIGCN-w/o-Temp}: We removed the temporal correlated edges in the intra-modal refinement to investigate the impact of the temporally adjacent relation.
		
		\item \textbf{MIGCN-w/o-Seman}: We removed the semantic correlated edges in the intra-modal refinement to validate the influence of the semantic similarity relation.
		
		\item \textbf{MIGCN-w/o-Syntac}: We removed the syntactic dependency edges in the intra-modal refinement to demonstrate the effectiveness of the syntactic dependency.
		
		\item \textbf{MIGCN-w/o-Gate}: To explore the effect of the gated graph convolution in inter-modal refinement, we replaced it with the naive graph convolution in Eqn.($\ref{naive-inter}$).
	\end{itemize}
	
	Table~\ref{AblationStudy} presents the ablation study results, from which we could observe that: 1) MIGCN shows superiority over MIGCN-w/o-GCN and MIGCN-G3G, demonstrating the effectiveness of the multi-modal interaction graph in MIGCN. As for the two different multi-modal graph architectures, MIGCN performs the modality fusion after the intra-modal and inter-modal refinement, while MIGCN-G3G fuses the two modalities before the inter-modal refinement. The early modal fusion in MIGCN-G3G may affect the correlation capture among the video and sentence query in the temporal language localization task and thus result the inferior performance. 2) MIGCN outperforms MIGCN-w/o-Inter, indicating the effectiveness of the inter-modal refinement in our method. The reason may be that, the inter-modal refinement is the key of the temporal language localization task to capture the semantic correspondence between the video and sentence query, and thus is pivotal to ensure the performance. 3) MIGCN surpasses MIGCN-w/o-Temp, MIGCN-w/o-Seman, and MIGCN-w/o-Syntac, suggesting the necessity of employing the intra-modal refinement in the task of temporal language localization in videos. The possible explanation is that the temporally adjacent relation and semantic correlation among video clips can promote the excavating of the sequential information and semantic clues among video clips, respectively, and hence boost the reasoning of the video. Moreover, the syntactic dependency among words is able to parse the sentence query and thus strengthen the representations learning of objects and actions in the sentence query. In a sense, these intra-modal relations are able to benefit the comprehensive understanding of the video and sentence query, which undoubtedly promote the inter-modal semantic correlation capture between the two modalities.	And 4) MIGCN outperforms MIGCN-w/o-Gate, which implies that the modality inherent information retained by the gate mechanism can contribute to the inter-modal refinement and performance improvement.
	
	\noindent \textbf{Ablation on Adaptive Context-Aware Localization.} To investigate the effect of the adaptive context-aware localization method, we devised the following derivations:
	\begin{itemize}
		\item \textbf{MIGCN-MP}: To check the impact of multi-scale fully connected layers in information maintaining, we employed the max-pooling operation to process the candidate moments with different lengths and then used a general fully connected layer to score the candidate moments and predict their offsets.
		
		\item \textbf{MIGCN-Sample}: Similar with~\cite{zhang2019cross}, we sampled the center vector of the candidate moment representation as its final representation and then scored the candidate moment and predicted its offsets same as MIGCN-MP.
		
		\item \textbf{MIGCN-w/o-Context}: To evaluate the effect of the context information, we scored the candidate moments and predicted the offsets without extending the candidate moments.
	\end{itemize}
	
	From Table~\ref{AblationStudy}, we observe that 1) a large performance decrease in both MIGCN-MP and MIGCN-Sample compared with MIGCN, which suggests that the information loss caused by MIGCN-MP and MIGCN-Sample can seriously deteriorate the performance. 2) MIGCN outperforms MIGCN-w/o-Context in most scenarios, confirming the importance of the context information in the candidate moment scoring and offset prediction. The underlying philosophy is that the context of candidate moment may convey the background or surrounding clues of the query moment, which can promote the localization accuracy. And 3) the performance improvement by context information on ActivityNet is more significant than that on Charades-STA. This phenomenon may be caused by the longer video duration of ActivityNet, where the video context are more informative for the localization.
	
	\begin{figure*}[t]
		\centering
		\subfloat[Example 1.]{
			\includegraphics[width=0.96\linewidth]{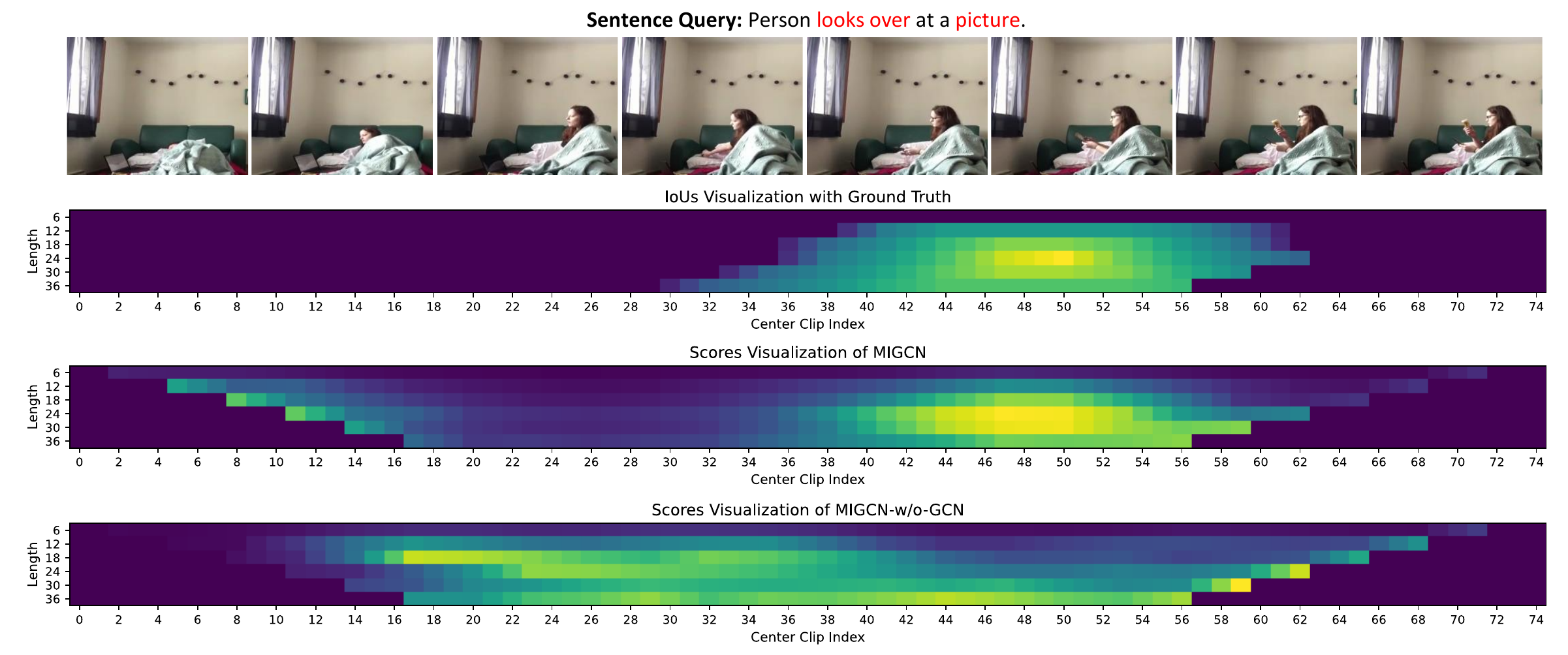}
			\label{score-example1}
		}
		\vspace{-0.5em}
		\subfloat[Example 2.]{
			\includegraphics[width=0.96\linewidth]{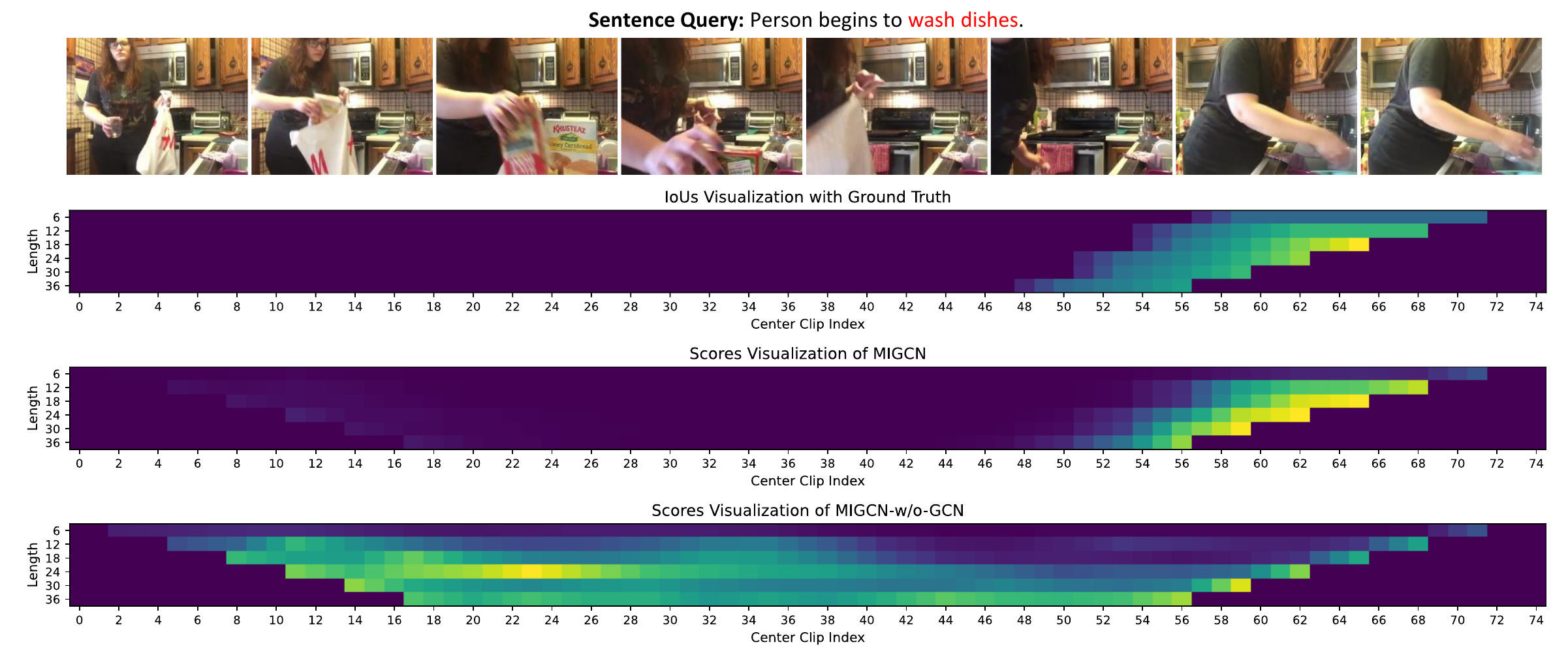}
			\label{score-example2}
		}
		\caption{Visualization of the candidate moment scores of MIGCN and MIGCN-w/o-GCN with two examples in Charades-STA dataset, where the IoUs of candidate moments with the ground truth are also provided for comparison. The horizontal and vertical coordinates stand for the center clip index and length of candidate moment, respectively. Scores in each row are smoothed by linear interpolation. Lighter colors indicate higher candidate moment scores.}
		\label{score-example}
	\end{figure*}
	
	\subsection{Efficiency Analysis}
	To gain the comprehensive understanding of our proposed MIGCN, we also analyzed its efficiency. We calculated the computation time~(Time)~\cite{Yuan_2019}, the frames per second~(FPS)~\cite{chen-etal-2018-temporally}, the floating point operations~(FLOPs)~\cite{TIP2020trg},  and the number of parameters~(Params)~\cite{TIP2020trg} of different methods. In particular, Time is defined as the average time to localize one sentence query in the video. FPS is calculated as dividing the number of frames in the video by its computation time. FLOPs is defined as the average floating point operations to localize one sentence query in the video. Params is the total number of learnable parameters in the model. Among these indicators, the model efficiency is mainly reflected by Time and FPS. FLOPs is the reference of the required computation of the method and calculated without considering the concrete implementation of model. Params is the reference of the required space. For a fair comparison, the feature extraction and data loading procedure are excluded in the Time, FPS and FLOPs calculation. It is worth noting that Time and FLOPs are not proportionally correlated. The underlying reason is that different methods are implemented with different parallel degree as they contain various network structures, \textit{e.g.}, linear layers and convolution layers.
	
	All the experiments are conducted on a single NVIDIA GeForce RTX 2080 Ti GPU. For a fair comparison, we only compare our MIGCN with those methods that have code and hyper-parameters publicly released. Except for the comparison, to show the highest efficiency of our method, we also display the results of MIGCN on Charades-STA with I3D feature, which has the minimum feature dimension compared with C3D and Two-Stream features.
	
	From Table~\ref{Efficiency}, we observe that: 1) Among all the proposal-based methods (\textit{i.e.}, CTRL, MAC, DRN, 2D-TAN and MIGCN), the Time of MIGCN, 2D-TAN and DRN are much shorter than that of CTRL and MAC. This may due to the fact that the former methods localize one moment with a single run of the model, while the latter ones require multiple times of model running to score all the candidate moments and then localize the target moment, which is time-consuming. 2) Compared with other single run models (\textit{i.e.}, 2D-TAN and DRN), our MIGCN spends less Time and has lower FLOPs and Params. 3) Although MIGCN is based on the dense proposal generation and ranking, the Time and FPS of MIGCN on Charades-STA with C3D are comparable with the proposal-free reinforcement learning based TSP-PRL. Therefore, we can draw the conclusion that our MIGCN has not only the promising performance but also the superior efficiency.
	
	\begin{figure*}[t]
		\centering
		\subfloat[Example 1.]{
			\includegraphics[width=0.96\linewidth]{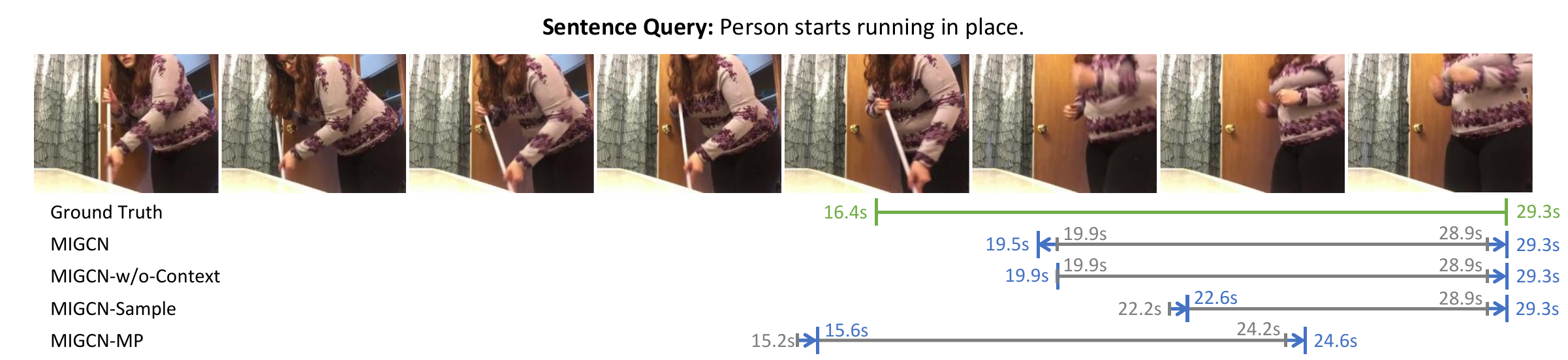}
			\label{ct-example1}
		}
		\vspace{-0.5em}
		\subfloat[Example 2.]{
			\includegraphics[width=0.96\linewidth]{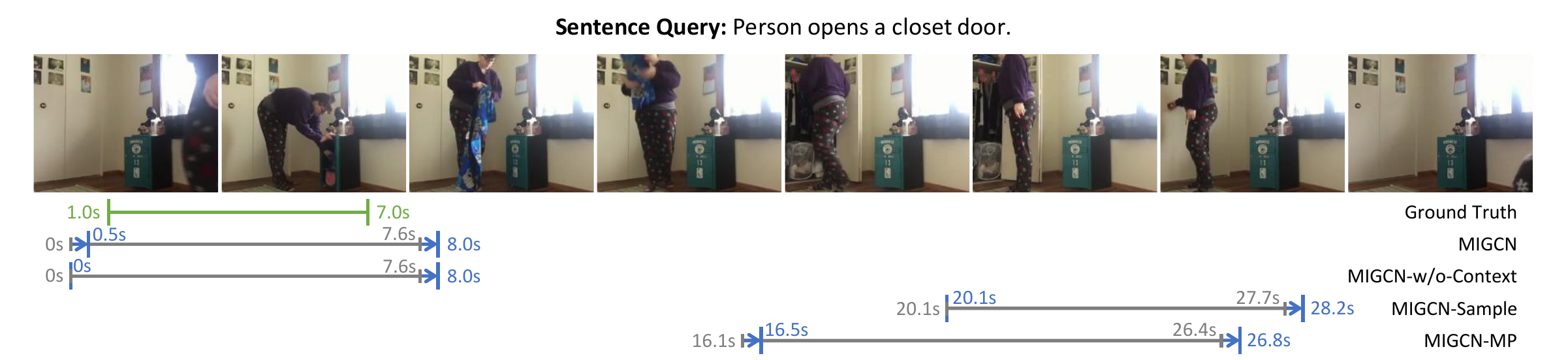}
			\label{ct-example2}
		}
		\caption{Prediction examples of MIGCN, MIGCN-w/o-Context, MIGCN-Sample and MIGCN-MP on Charades-STA dataset. The annotations in green color represent the ground truth moments. Those in gray color denote the retrieved coarse candidate moments without boundary adjustment, while those in blue represent the predicted locations after boundary adjustment.}
		\label{ct-example}
	\end{figure*}
	
	\subsection{Result Visualization}
	In order to gain more deep insights regarding the effects of intra-modal relations as well as inter-modal interactions, we visualized the scores of candidate moments predicted by MIGCN and MIGCN-w/o-GCN with examples in Figure~\ref{score-example}. In Figure~\ref{score-example}\subref{score-example1}, given the sentence query ``Person looks over at a picture'', MIGCN-w/o-GCN highly scores all the candidates that contain the action of a person looking over at some objects, such as laptop, curtain and bottle. Beyond that, MIGCN gives the high score to the candidate moment only when the person looks over at a picture, and thus obtains a more similar score distribution with the ground truth. This may be attributed to the fact that MIGCN not only understands the more concrete action ``looks over at the picture'' due to the intra-modal relation, \textit{i.e.}, syntactic dependency, but also comprehensively considers the inter-modal interactions between the sentence query and video, and giving more rational candidate moment scores. As to the slightly high score predictions of MIGCN at the beginning of the video, one possible explanation is that these candidate moments contain the action that the person looks over at the laptop, which is rather indistinguishable with the picture. Similarly, in Figure~\ref{score-example}\subref{score-example2}, MIGCN gives high scores to the candidates where the person ``wash dishes'', which is more accurate than MIGCN-w/o-GCN, since the latter only sees the ``dishes'' in the video while neglecting the key action ``wash''.
	
	Moreover, we provided some prediction results of MIGCN, MIGCN-w/o-Context, MIGCN-Sample and MIGCN-MP to intuitively show the effect of the adaptive context-aware localization method in Figure~\ref{ct-example}. As we could see from the two examples, with the multi-scale fully connected layers, MIGCN selects the more ideal coarse candidate moments than MIGCN-Sample and MIGCN-MP, indicating that compared with the sampling and pooling methods, the multi-scale fully connected layers of MIGCN are able to rank the variable-length candidate moments with less information loss and thus accurately locate the target moment. In addition, we observed that although MIGCN and MIGCN-w/o-Context select the same candidate moment, MIGCN obtains a higher IoU result with the more precise boundary adjustment. Similar observation can be found in the second example, which confirms the effect of the context information in the boundary adjustment.
	
	\section{Conclusion}
	In this work, we propose a multi-modal interaction graph convolutional network to tackle the task of temporal language localization in videos, which promotes the comprehensive understanding of the video and sentence query and facilitates their semantic correspondence capture with both the intra-modal relation and inter-modal interaction modeling. Moreover, we devise an adaptive context-aware localization method to calculate the ranking scores and boundary offsets of the coarse candidate moments, which considers the context and retains the information of candidate moments as much as possible with the multi-scale fully connected layers. Extensive experiments on Charades-STA and ActivityNet datasets verify the superior effectiveness and efficiency of our method compared with the state-of-the-art methods. Furthermore, the ablation study confirms the virtues of intra-modal relations, inter-modal interactions and adaptive context-aware localization method on this task. As for the future work, on the one hand, we plan to investigate the effect of visual and linguistic representations on vision-language task. On the other hand, we would enhance the generalization of the model to adapt to various datasets (\textit{i.e.}, TACoS~\cite{tacos}).

	
	%



	\ifCLASSOPTIONcaptionsoff
	\newpage
	\fi

	
	
	\bibliographystyle{IEEEtran}
	\bibliography{IEEEabrv,mybib}
	%
	%
	%
	
	%
	
	\begin{IEEEbiography}[{\includegraphics[width=1in,height=1.25in,clip,keepaspectratio]{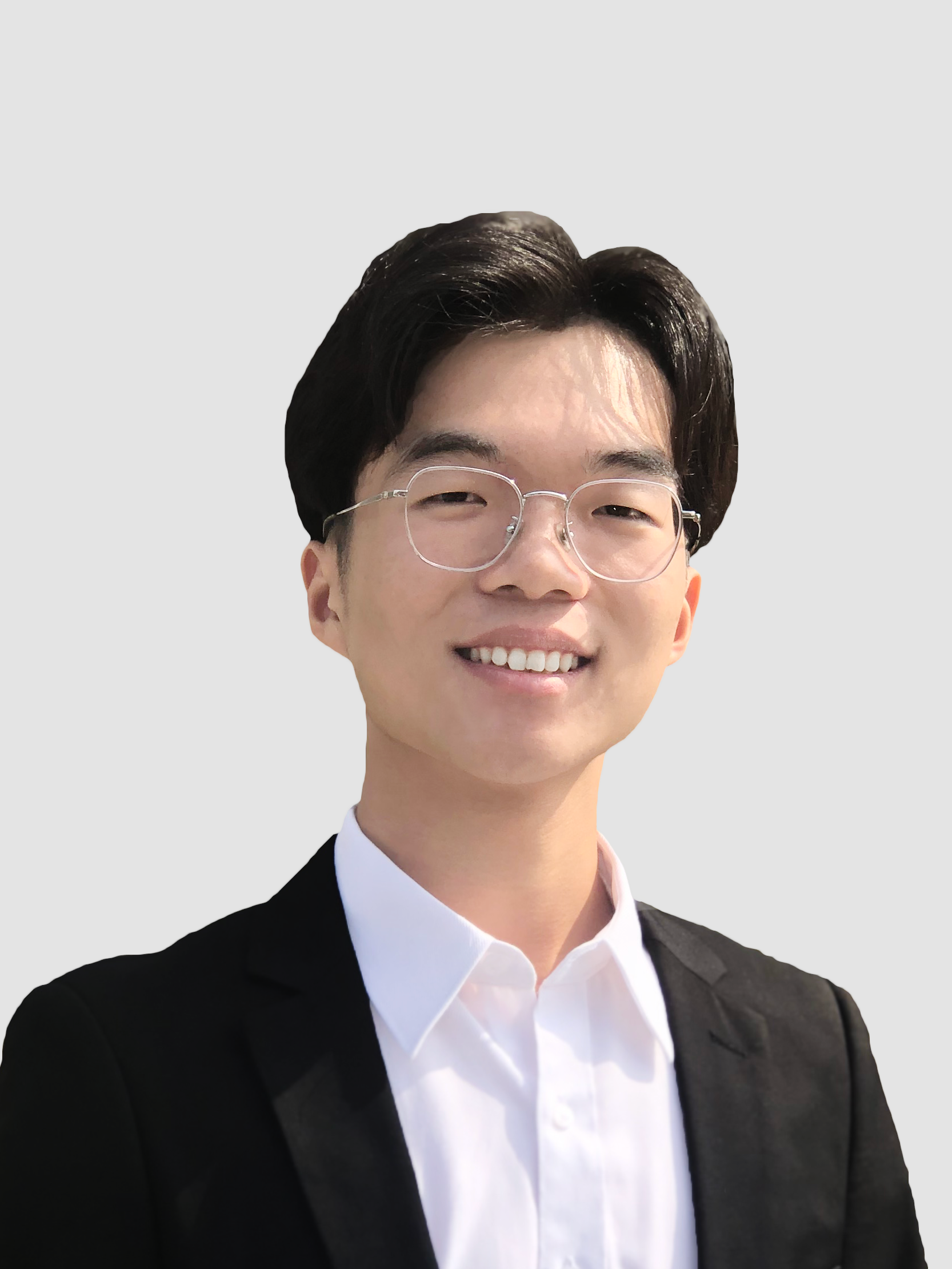}}]{Zongmeng Zhang}
		received the B.E. degree from Shandong University in 2021. He is currently pursuing the master's degree in University of Science and Technology of China. His research interests include computer vision and multimedia computing.
	\end{IEEEbiography}
	
	\begin{IEEEbiography}[{\includegraphics[width=1in,height=1.25in,clip,keepaspectratio]{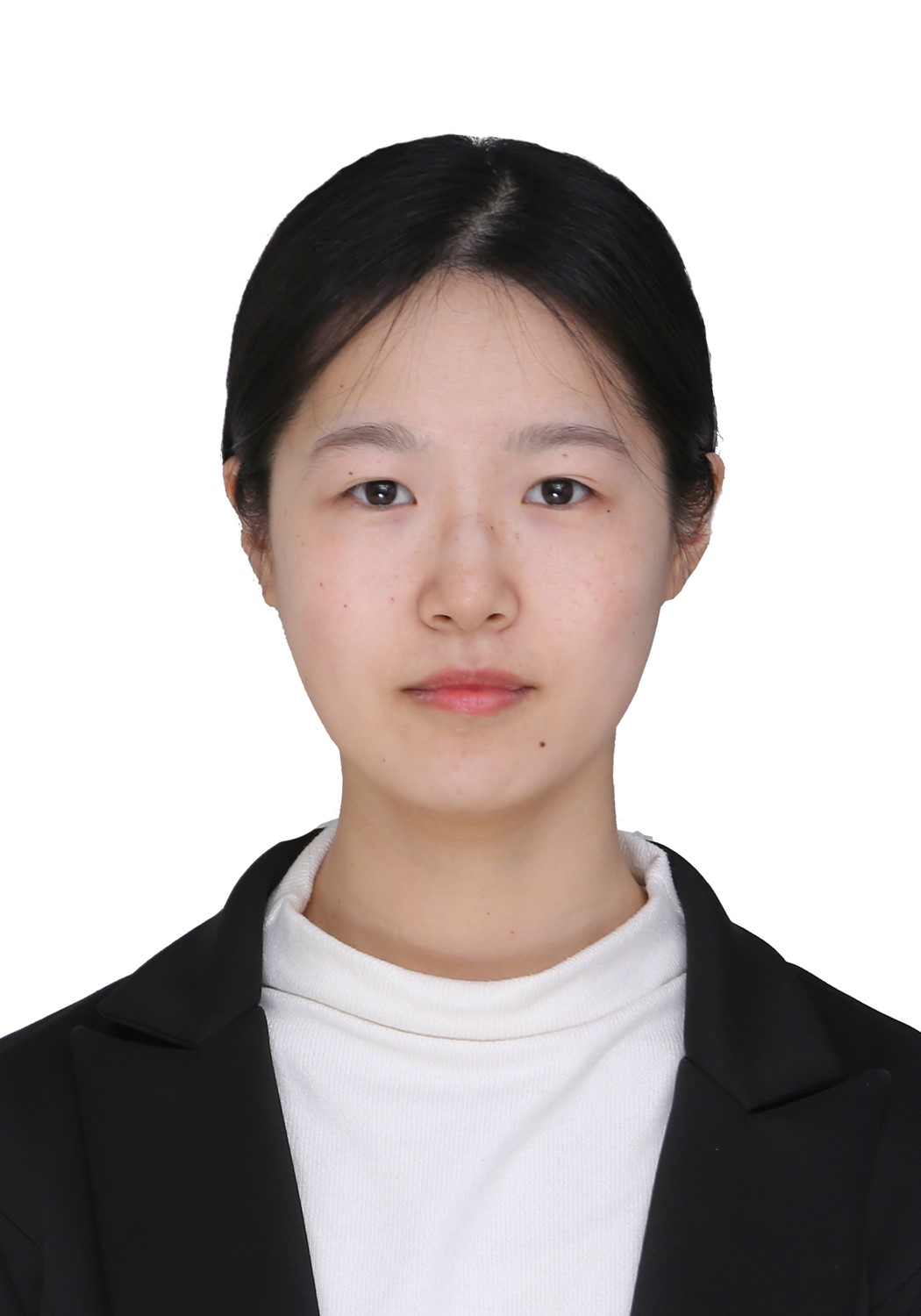}}]{Xianjing Han}
		received the B.E. degree from Northeastern University, China, in 2017. She is currently pursuing the Ph.D. degree with the School of Computer Science and Technology, Shandong University, under the supervision of Professor Liqiang Nie and Professor Xuemeng Song. Her research interests include multimedia computing and computer vision.
	\end{IEEEbiography}
	
	\begin{IEEEbiography}[{\includegraphics[width=1in,height=1.25in,clip,keepaspectratio]{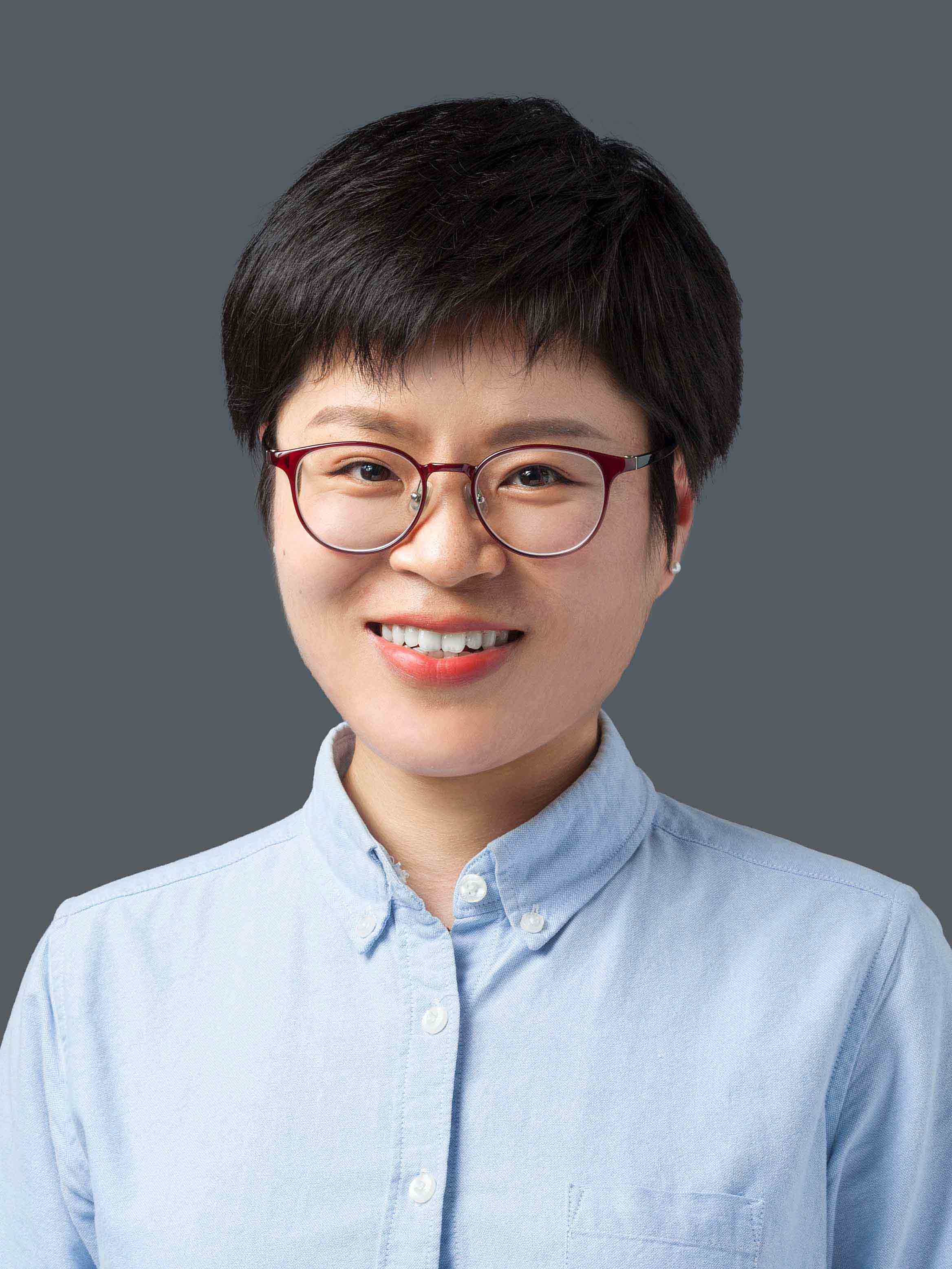}}]{Xuemeng Song}
		received the B.E. degree from University of Science and Technology of China in 2012, and the Ph.D. degree from the School of Computing, National University of Singapore in 2016. She is currently an associate professor of Shandong University, Jinan, China. Her research interests include the information retrieval and social network analysis. She has published several papers in the top venues, such as ACM SIGIR, MM and TOIS. In addition, she has served as reviewers for many top conferences and journals.
	\end{IEEEbiography}
	
	\begin{IEEEbiography}[{\includegraphics[width=1in,height=1.25in,clip,keepaspectratio]{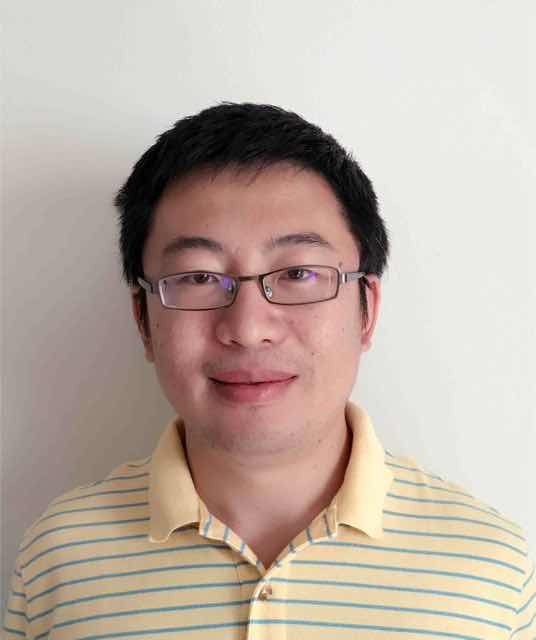}}]{Yan Yan}
		is currently a Gladwin Development Chair Assistant Professor in the Department of Computer Science at Illinois Institute of Technology. He was an assistant professor at the Texas State University, a research fellow at the University of Michigan and the University of Trento. He received his Ph.D. in Computer Science at the University of Trento. His research interests include computer vision, machine learning and multimedia.
	\end{IEEEbiography}
	
	\begin{IEEEbiography}[{\includegraphics[width=1in,height=1.25in,clip,keepaspectratio]{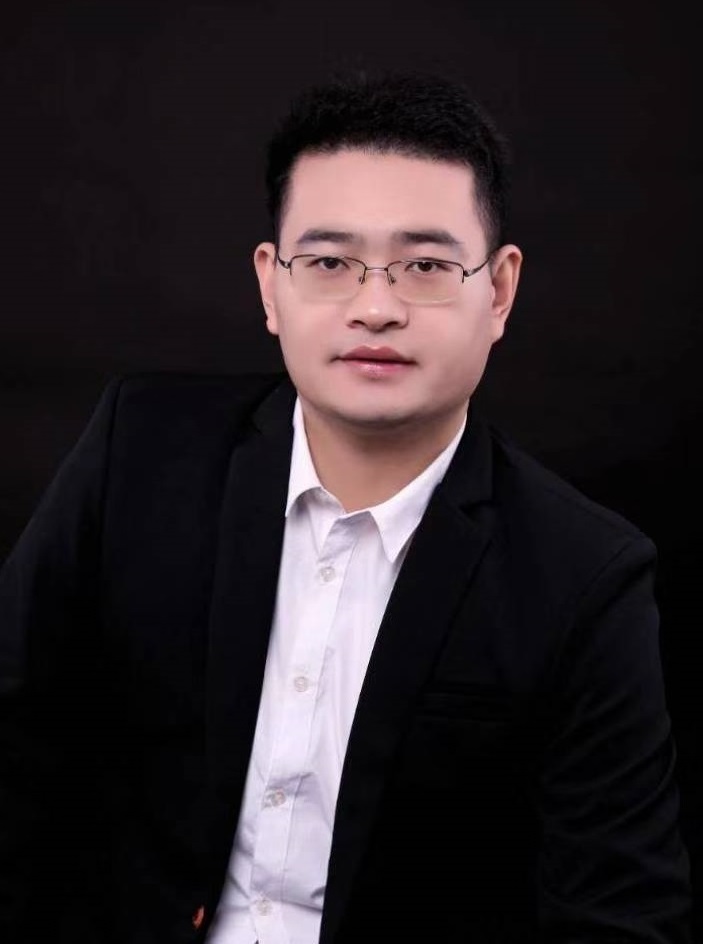}}]{Liqiang Nie}
		is currently a professor with the School of Computer Science and Technology, Shandong University. Meanwhile, he is the adjunct dean with the Shandong AI institute. He received his B.Eng. and Ph.D. degree from Xi'an Jiaotong University in July 2009 and National University of Singapore (NUS) in 2013, respectively. After PhD, Dr. Nie continued his research in NUS as a research follow for more than three years. His research interests lie primarily in multimedia computing and information retrieval. Dr. Nie has co-/authored more than 140 papers, received more than 7,800 Google Scholar citations as of Jun. 2019. He is an AE of Information Science, an area chair of ACM MM 2018, a special session chair of PCM 2018, a PC chair of ICIMCS 2017. Meanwhile, he is supported by the program of ``Thousand Youth Talents Plan 2016'', ``Qilu Scholar 2016'', and ``The Shandong Province Science Fund for Distinguished Young Scholars 2018''. In 2017, he co-founded ``Qilu Intelligent Media Forum''.
	\end{IEEEbiography}
	
	
	
	

\end{document}